\documentclass[runningheads]{llncs}
\usepackage[T1]{fontenc}
\usepackage{graphicx}
\usepackage{xcolor}
\usepackage{mwe} 
\usepackage{boldline,tabularx}
\usepackage{amsmath,amssymb,bm,bbm,mathtools}
\DeclareMathOperator*{\argmax}{argmax}
\DeclareMathOperator*{\argmin}{argmin}
\usepackage{paralist}
\usepackage{hyperref}
\usepackage[misc]{ifsym}

\usepackage{caption} 
\begin{document}
\title{MRF-UNets: Searching UNet with Markov Random Fields}
%
%
\author{Zifu Wang \Letter
\and
Matthew B. Blaschko}
\authorrunning{Z. Wang and M. B. Blaschko}
%
\institute{ESAT-PSI, KU Leuven, Leuven, Belgium \\
\email{zifu.wang}@kuleuven.be}
\maketitle              
\begin{abstract}
UNet \cite{U-NetRonnebergerMICCAI2015} is widely used in semantic segmentation due to its simplicity and effectiveness. However, its manually-designed architecture is applied to a large number of problem settings, either with no architecture optimizations, or with manual tuning, which is time consuming and can be sub-optimal. In this work, firstly, we propose Markov Random Field Neural Architecture Search (MRF-NAS) that extends and improves the recent Adaptive and Optimal Network Width Search (AOWS) method \cite{AOWSBermanCVPR2020} with (i) a more general MRF framework (ii) diverse M-best loopy inference (iii) differentiable parameter learning. This provides the necessary NAS framework to efficiently explore network architectures that induce loopy inference graphs, including loops that arise from skip connections. With UNet as the backbone, we find an architecture, MRF-UNet, that shows several interesting characteristics. Secondly, through the lens of these characteristics, we identify the sub-optimality of the original UNet architecture and further improve our results with MRF-UNetV2. Experiments show that our MRF-UNets significantly outperform several benchmarks on three aerial image datasets and two medical image datasets while maintaining low computational costs. The code is available at: \href{https://github.com/zifuwanggg/MRF-UNets}{https://github.com/zifuwanggg/MRF-UNets}.
\end{abstract}

\begin{keywords}
Neural Architecture Search, Probabilistic Graphical Models, Semantic Segmentation
\end{keywords}

\section{Introduction}
Neural architecture search (NAS) has greatly improved the performance on various vision tasks, for example, classification \cite{AOWSBermanCVPR2020,FBNetV3DaiCVPR2021,OnWanICLR2022}, object detection \cite{Auto-FPNXuICCV2019,OPANASLiangCVPR2021,NCPDingICLR2022} and semantic segmentation \cite{NAS-UnetWengAccess2019,MS-NASYanMICCAI2020,BiX-NASWangMICCAI2021,AdwU-NetHuangMIDL2022} via automating the architecture design process. UNet \cite{U-NetRonnebergerMICCAI2015} is widely adopted to a large number of problem settings such as aerial \cite{DeepGlobeDemirCVPRWorkshop2018,AdaptiveSrivastavaBMVC2019} and medical image segmentation \cite{PROMISE12LitjensMIA2014,OptimizingTMI2020,CHAOSKavurMIA2021} due to its simplicity and effectiveness, but either with no architecture optimization, or with simple manual tuning. A natural question is, can we improve its manually-designed architecture with NAS?

AOWS \cite{AOWSBermanCVPR2020} is a resource-aware NAS method and it is able to find effective architectures that strictly satisfy resource constraints, e.g. latency or the number of floating-point operations (FLOPs). The main idea of AOWS is to model the search problem as parameter learning and maximum a posteriori (MAP) inference over a Markov Random Field (MRF). However, the main limitation of AOWS is the adoption of Viterbi inference. As a result, the approach is only applicable to simple tree-structured graphs where the architecture cannot have skip connections \cite{ResNetHeCVPR2016}. Skip connections are widely adopted in modern neural networks as they can ease the training of deep models via shortening effective paths \cite{ResidualVeitNeurIPS2016}; skip connections have also played an important role in the success of semantic segmentation where an encoder and a decoder are connected by long skip connections to aggregate features at different levels, e.g. UNet.

Besides, MAP assignment over a weight-sharing network is usually sub-optimal due to the discrepancy between the one-shot super-network and stand-alone child-networks \cite{NASYangICLR2020,EvaluatingYuICLR2020}. Restricting the search to a single MAP solution also results in the search method having high variance \cite{DARTSLiuICLR2019}, so one usually needs to repeat the search process with different random seeds and hyper-parameters. Furthermore, parameter learning of AOWS is non-differentiable, and this disconnects AOWS from recent advances in the differentiable NAS community \cite{DARTSLiuICLR2019,FairDARTSChuECCV2020,BetaDARTSYeCVPR2022}, despite its advantages in efficient inference. 

The contributions of this paper are twofold. Firstly, we propose MRF-NAS, which extends and improves AOWS with 
\begin{inparaenum}[(i)]
\item a more general framework that shows close connections with other NAS approaches and yields better representation capability,
\item loopy inference algorithms so we can apply it to more complex search spaces,
\item diverse M-best inference instead of a single MAP assignment to reduce the search variance and to improve search results and
\item a novel differentiable parameter learning approach with Gibbs sampling and Long-Short-Burnin-Scheme (LSBS) to save on computational cost.
\end{inparaenum}
With UNet as the backbone, we find an architecture, MRF-UNet, that shows several interesting characteristics. Secondly, through the lens of these characteristics, we identify the sub-optimality of the original UNet architecture and further improve our results with MRF-UNetV2. We show the effectiveness of our approach on three aerial image datasets: DeepGlobe Land, Road and Building \cite{DeepGlobeDemirCVPRWorkshop2018} and two medical image datasets: CHAOS \cite{CHAOSKavurMIA2021} and PROMISE12 \cite{PROMISE12LitjensMIA2014}. Compared with the benchmarks, our MRF-UNets achieve superior performance while maintaining low computational cost.

\section{Related Works}
Neural architecture search (NAS) \cite{NASZophICLR2017,DARTSLiuICLR2019} is a technique for automating the design process of neural network architectures. The early attempt \cite{NASZophICLR2017} trains a RNN with reinforcement learning and costs thousands of GPU hours. In order to reduce the search cost, one usually uses some proxy to infer an architecture's performance without training it from scratch. For example, the significance of learnable architecture parameters \cite{DARTSLiuICLR2019,FairDARTSChuECCV2020,BetaDARTSYeCVPR2022,NAS-UnetWengAccess2019,MS-NASYanMICCAI2020,DiNTSHeCVPR2021,BiX-NASWangMICCAI2021,AdwU-NetHuangMIDL2022} or validation accuracy from a super-network with shared weights \cite{UnderstandingBenderICML2018,AOWSBermanCVPR2020,SingleGuoECCV2020,FairNASChuICCV2021}. 

Resource-aware NAS \cite{AutoSlimYuNeurIPSWorkshop2019,AOWSBermanCVPR2020,BigNASYuECCV2020,FBNetV3DaiCVPR2021} focuses on architectures that achieve good performance while satisfy resource targets such as FLOPs or latency. FBNet \cite{FBNetV3DaiCVPR2021} inserts a differentiable latency term into the loss function to penalize networks that consume high latency. However, the found architectures are not guaranteed to strictly satisfy the constraints. AutoSlim \cite{AutoSlimYuNeurIPSWorkshop2019} trains a slimmable network \cite{SNNYuICLR2019,UniversallyYuICCV2019} as the super-network and applies a greedy heuristics to search for channel configurations under different FLOPs targets. AOWS \cite{AOWSBermanCVPR2020} models resource-aware NAS as a constrained optimization problem which can then be solved via inference over a chain-structured MRF. Nevertheless, their method can only be applied to simple search spaces which do not include skip connections.

\section{Preliminaries}
\subsection{Markov Random Field} 
For an arbitrary integer $n$, let $[n]$ be shorthand for $\{1, 2, ..., n\}$. We have a set of discrete variables $\bm{x} = \{x_i | i\in [n]\}$ and each $x_i$ takes value in a finite label set $X_i = \{x_i^j | j\in [k_i] \}$. For a set $S \subseteq [n]$, we use $x_S$ to denote $\{x_i | i\in S\}$, and $X_S = \bigtimes_{i\in S} X_i$, where $\bigtimes$ is the cartesian product. 

A Markov Random Field (MRF) is an undirected graph $G=(V,E)$ over these variables, and equipped with a set of factors $\Phi=\{\phi_S|S \subseteq [n]\}$ where $\phi_S: X_S \rightarrow \mathbb{R}$, such that $V=[n]$ and an edge $e_{i,j} \in E$ when there exists some $\phi_S\in \Phi$ and $\{i,j\} \subseteq S$ \cite{PGMKoller2009}. It is common to employ a pairwise MRF where $\Phi=\{\phi_S \big| S \subseteq [n] \text{ and } |S| \leq 2\}$. A set of factors $\Phi$ explicitly defines a probabilistic distribution $\mathbb{P}_{\Phi}(\bm{x}) = \frac{1}{Z} \exp \Big( \sum_{S}{\phi_S(x_S)} \Big)$ where $Z$ is the normalizing constant. The goal of MAP inference is to find an assignment $\bm{x}^*$ so as to maximize a real-valued energy function $\mathcal{E}(\bm{x})$:
\begin{equation}
\bm{x}^*=\argmax_{\bm{x} \in X_V} \mathcal{E}(\bm{x}) = \argmax_{\bm{x} \in X_V} \exp \Big( \sum_{S}{\phi_S(x_S)} \Big).
\end{equation}

\subsection{Diverse M-best Inference} \label{sec:diverse}
In MRFs, there exist optimization error (approximate inference), approximation error (limitations of the model, e.g. a pairwise MRF can only represent pairwise interactions), and estimation error (factors are learnt from a finite dataset). In the context of NAS, in order to reduce the cost of searching, we often resort to proxies \cite{UnderstandingBenderICML2018,DARTSLiuICLR2019} on a weight-sharing network and they can be inaccurate. Instead of giving all our hope to a single MAP solution, diverse M-best inference \cite{DiverseMBestBatraECCV2012} aims to find a diverse set of highly probable solutions.

Given some dissimilarity function $\Delta(\bm{x}^p,\bm{x}^{q})$ between two solutions and a dissimilarity target $k^{q}$, we denote $\bm{x}^1$ as the MAP,  $\bm{x}^2$ the second-best solution and so on until $\bm{x}^m$ the $m$th-best solution. Then for each $2 \leq p \leq m$, we have the following constrained optimization problem
\begin{align}
\label{eq:diverse}
\bm{x}^{p} &=\argmax_{\bm{x} \in X_V} \mathcal{E}(\bm{x}) \qquad \\ 
&\text{s.t.} \; \Delta(\bm{x}^{p},\bm{x}^{q})\geq k^{q} \quad \text{for } q=1,...,p-1.
\end{align}
Therefore, we are interested in a diverse set of solutions $\{\bm{x}^{1},...,\bm{x}^p,...,\bm{x}^{m}\}$ such that each $\bm{x}^p$ maximizes the energy function, and is at least $k^{q}$-units away from each of the $p-1$ previously found solutions. If we consider a pairwise MRF and choose Hamming distance as the dissimilarity function, we can turn Eq. (\ref{eq:diverse}) into a new MAP problem such that pairwise factors remain the same and unary factors become $\phi_i^p(x_i^j)= \phi_i(x_i^j) - \sum_{q=1}^{p-1}\lambda^q \cdot \mathbbm{1}(x_i^{\boldsymbol{\cdot}}{}^q = x_i^j)$ where $\lambda^q$ is the Lagrange multiplier and $x_i^{\boldsymbol{\cdot}}{}^q$ is the assignment of $x_i$ in the q-th solution \cite{DiverseMBestBatraECCV2012}.

\subsection{AOWS} \label{sub:AOWS}
Having introduced the notations for MRFs, here we illustrate how AOWS models the NAS problem as a MRF. In NAS, there is a neural network $N$ that has $n$ choice nodes, i.e. $\bm{x} = \{x_i | i\in [n]\}$, and each node $x_i$ can take some value from a label set $X_i$, e.g. kernel size = 3 or 5. Therefore, we can use $\bm{x}$ to represent the architecture of a neural network. Let $N(\bm{x})$ be a neural network whose architecture is $\bm{x}$. Given some task-specific performance measurement $\mathcal{M}$, e.g. classification accuracy, and a resource measurement $\mathcal{R}$, e.g. latency or FLOPs, resource-aware NAS can be represented as a constrained optimization problem
\begin{equation}
\max_{\bm{x}} \mathcal{M}(N(\bm{x})) \quad \text{s.t.} \; \mathcal{R}(N(\bm{x})) \leq R_T 
\end{equation}
where $R_T$ is the resource target. Consider the following Lagrangian relaxation of the problem
\begin{equation}
\label{eq:minmax}
\min_{\gamma}\max_{\bm{x}} \mathcal{M}(N(\bm{x})) + \gamma(\mathcal{R}(N(\bm{x})) - R_T )
\end{equation}
with $\gamma$ a Lagrange multiplier. If the inner maximization problem can be solved effienciently, then the minimization problem in Eq. (\ref{eq:minmax}) can be solved by binary search over $\gamma$ since the objective is concave in $\gamma$ \cite{AdaptiveSrivastavaBMVC2019,AOWSBermanCVPR2020}.

The key idea of AOWS \cite{AOWSBermanCVPR2020} is to model Eq. (\ref{eq:minmax}) as parameter learning and MAP inference over a pairwise MRF such that $\mathcal{M}(N(\bm{x})) = \sum_i{\phi_i}$ and $\mathcal{R}(N(\bm{x})) = \sum_{i,j}{\phi_{i,j}}$. For $\mathcal{M}(N(\bm{x}))$, they assume $\phi_i(x_i^j) = -\frac{1}{|T_{i,j}|}\sum_{t}l(\bm{w}|x_i^{(t)}=j)$ where $l(\bm{w}|x_i^{(t)}=j)$ is the training loss when $x_i=j$ is sampled at iteration $t$, and $|T_{i,j}|$ is the total number of times $x_i^j$ is sampled. For $\mathcal{R}(N(\bm{x}))$, many resource models have a pairwise form. For example, FLOPs can be calculated exactly as pairwise sums; latency is usually modeled as a pairwise model due to sequential execution of the forward pass. Once these factors are known, the inner maximization problem can be solved efficiently via Viterbi inference.

\section{MRF-NAS}
Here we generalize the idea of AOWS \cite{AOWSBermanCVPR2020} to a broader setting. We assume that there exists some non-decreasing mapping $\mathcal{F}: \mathbb{R}\rightarrow \mathbb{R}$ such that
\begin{equation}
\label{eq:map}
\mathcal{M}(N(\bm{x}))=\mathcal{F}(\mathcal{E}(\bm{x}))
\end{equation}
Therefore, we extend their framework and no longer require $\mathcal{M}(N(\bm{x})) = \mathcal{E}(\bm{x})$, but let $\mathbb{P}_{\Phi}$ be defined by a set of factors $\Phi$ such that $\mathbb{P}_{\Phi}(\bm{x}_1) \geq  \mathbb{P}_{\Phi}(\bm{x}_2) \Rightarrow \mathcal{M}(N(\bm{x}_1)) \geq \mathcal{M}(N(\bm{x}_2))$. Then NAS becomes MAP inference over a set of properly defined factors
\begin{equation}
\bm{x}^*=\argmax_{\bm{x}} \mathcal{M}(N(\bm{x})) = \argmax_{\bm{x}} \mathcal{E}(\bm{x}).
\end{equation}
For resource-aware NAS, we follow \cite{AOWSBermanCVPR2020} to introduce another set of factors
\begin{equation}
    \mathcal{R}(N(\bm{x})) = \mathcal{E}'(\bm{x}) = \exp \Big( \sum_{i \in V}\phi_i'(x_i) + \sum_{(i,j) \in E}\phi_{i,j}'(x_i,x_j) \Big)
\end{equation}
and combine these two energy functions as in Eq. (\ref{eq:minmax}). As discussed in section \ref{sub:AOWS}, many resource models can be represented exactly or approximately as a pairwise model. Following \cite{AOWSBermanCVPR2020}, we focus on latency. We can populate each element $\phi_i'$ and $\phi_{i,j}'$ through profiling the entire network on some target hardware and solving a system of linear equations.

Many existing methods show close connections to our formulation. For example, one-shot methods with weight sharing \cite{UnderstandingBenderICML2018,SingleGuoECCV2020,FairNASChuICCV2021} define a single factor $\phi_V$ whose scope includes all nodes in the graph where $\phi_V(\bm{x})$ is the validation accuracy of the super-network evaluated with architecture $\bm{x}$. Their formulation imposes no factorization, and therefore the cardinality of $\phi_V(\bm{x})$ grows exponentially in the order of $n$, the number of nodes in the graph, which makes MAP inference impossible. On the contrary, AOWS \cite{AOWSBermanCVPR2020} and differentiable NAS approaches \cite{DARTSLiuICLR2019,FairDARTSChuECCV2020,BetaDARTSYeCVPR2022} introduce a set of unary factors $\Phi=\{\phi_i|i \in V\}$ such that in AOWS, $\phi_i(x_i)$ is the averaged losses, and in differentiable NAS approaches, $\phi_i(x_i)$ is the learnable architecture parameter. With no higher order interaction, MAP inference deteriorates to marginal maximization whose solution can be derived easily. However, this model imposes strong local independence and greatly limits the representation capability of the underlying graphical model.

Since we model the resource measurement as a pairwise model, for the ease of joint inference in Eq. (\ref{eq:minmax}), we also consider $\mathcal{E}(\bm{x})$ to be pairwise 
\begin{equation}
\label{eq:pairwise}
\mathcal{E}(\bm{x}) = \exp \Big( \sum_{i \in V}\phi_i(x_i) + \sum_{(i,j) \in E}\phi_{i,j}(x_i,x_j) \Big).
\end{equation}
Moreover, compared with methods that only use unary terms, our pairwise model imposes weaker local independence and has more representation power. Although the inclusion of pairwise terms increases the number of learnable factors from $O(|X|)$ to $O(|X|^2)$, where $|X|$ is the cardinality of factors and 
is usually less than 20, the added overhead is negligible compared with the number in learnable parameters of modern neural networks.

\subsection{Diverse M-best Loopy Inference} 
The main limitation of AOWS \cite{AOWSBermanCVPR2020} is the adoption of Viterbi inference, which is only applicable to simple chain graphs such as MobileNetV1 \cite{MobileNetsHowardarXiv2017}. When the computational graph forms loops, i.e.\ it includes skip connections \cite{ResNetHeCVPR2016}, we need to resort to loopy inference algorithms. Exact inference over a loopy graph can be very expensive, especially when the graph is densely connected. In the worst case, the complexity of exact inference can be exponential in $n$ when the graph is fully connected. Therefore, sometimes we can only hope for approximate solutions.  However, we find in practice that for realistic architectures, fast approximate inference yields excellent performance on par with exact inference. The difference between exact and approximate inference will be discussed in more detail in section \ref{sec:inference}.

Furthermore, MAP assignment on a weight-sharing network is usually of poor quality, but we can still find architectures that achieve good performance by examining other top solutions \cite{NASYangICLR2020,EvaluatingYuICLR2020}. Therefore, instead of a single MAP assignment $\bm{x}^*$, we use diverse M-best inference \cite{DiverseMBestBatraECCV2012} to find a set of diverse solutions $\{\bm{x}^1,...,\bm{x}^m\}$ so as to reduce the variance in the search phase.  

Diverse M-best inference requires a set of balanced dissimilarity constraints, each with an associated Lagrange multiplier. In Eq. (\ref{eq:diverse}), it is crucial to choose a dissimilarity target $k^q$, and rather than searching via the diversity constraints, we can directly perform model selection on the Lagrange multiplier $\lambda^q$ \cite{DiverseMBestBatraECCV2012}. We find that the absolute value of $\phi_i$ can be very different across factors. Instead of using a single scalar value $\lambda^q$ for all $i$, we set it to be a vector $\bm{\lambda}^q = (\lambda_1^q, ..., \lambda_i^q, ..., \lambda_n^q)$ such that
\begin{equation} \label{eq:L}
\lambda_i^q = \frac{\max_j \phi_i^q(x_i^j) - \min_j \phi_i^q(x_i^j)}{L},
\end{equation}
where $\phi_i^q(\cdot)$ is the modified unary factor. Then we can tune $L$ instead.

\subsection{Differentiable Parameter Learning}
In the previous sections, we have discussed how to find optimal solutions if the factors in MRF are already known. Here we propose a differentiable approach to learn these factors so as to close the gap between AOWS and other differentiable NAS approaches. Following the formulation in \cite{NADSArdywibowoICML2020}, the goal of differentiable NAS is to maximize the following objective
\begin{equation}
\label{eq:obj}
-\mathbb{E}_{\bm{x} \sim \mathbb{P}_{\Phi}(\bm{x})}[l (\bm{w}^*|\bm{x})] \quad \text{s.t.} \; \bm{w}^*=\argmin_{\bm{w}}l(\bm{w}|\bm{x}),
\end{equation}
where $l(\cdot)$ is some loss function and $\bm{w}$ encodes connection weights of the neural network. In order to make Eq. (\ref{eq:obj}) differentiable, we can approximate it through Monte Carlo with $n_{mc}$ samples and use the Gumbel-Softmax reparameterization trick \cite{Gumbel-SoftmaxJangICLR2017} to smooth the discrete categorical distribution. 

\begin{figure}[t]
\begin{center}
\includegraphics[width=1\linewidth]{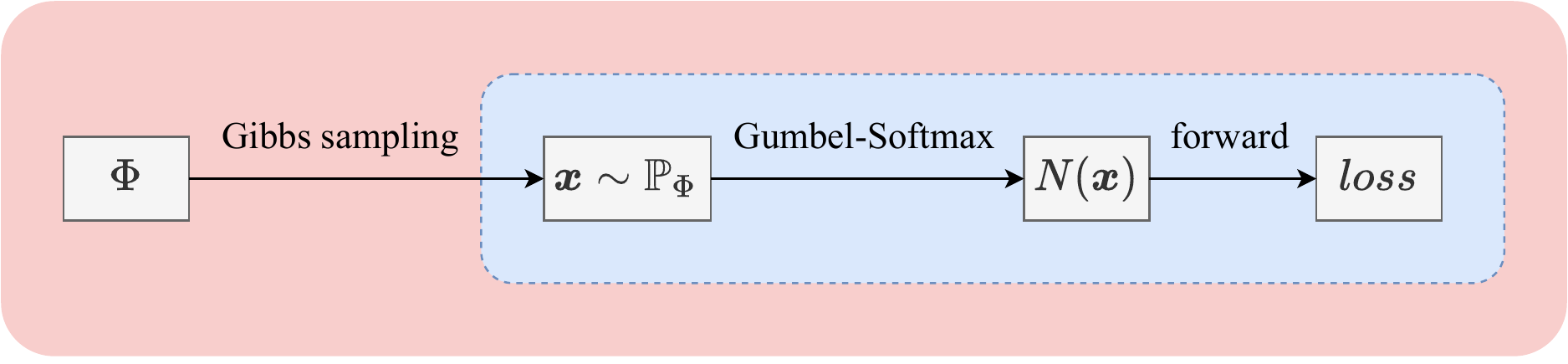}
\caption{\label{fig:learning}Workflow of our differentiable parameter learning approach. Vanilla differentiable NAS methods \cite{NADSArdywibowoICML2020} with Monte Carlo approximation and Gumbel-Softmax trick are shown in the blue rectangle. In our case, since the joint probability distribution $\mathbb{P}_{\Phi}(\bm{x})$ cannot be decomposed as the product of only unary terms, we need to perform an extra step of MCMC, e.g.\ Gibbs sampling.}
\end{center}
\end{figure}

However, there is one more caveat in the aforementioned approach: to sample from the joint probability distribution $\mathbb{P}_{\Phi}(\bm{x})$. When we only have unary factors such as in \cite{NADSArdywibowoICML2020}, sampling from $\mathbb{P}_{\Phi}(\bm{x})$ is the same as independently sampling from the marginal probability distribution $\mathbb{P}_{\phi_i}(x_i)$:
\begin{equation}
    \mathbb{P}_{\Phi}(\bm{x}) = \frac{1}{Z} \exp \Big( \sum_{i}{\phi_i(x_i)} \Big) = \prod_{i} \frac{1}{Z} \exp \Big( \phi_i(x_i) \Big) = \prod_{i} \mathbb{P}_{\phi_i}(x_i).
\end{equation}
When we have high-order interactions such as pairwise terms, sampling from the joint usually involves the use of Markov Chain Monte Carlo (MCMC) methods. Here we use Gibbs sampling for simplicity:
\begin{equation}
x_i^t \sim \mathbb{P}_{\Phi}(x_i|\bm{x}_{-i})
\end{equation}
where the distribution of $x_i$ at $t$-th iteration is determined by $\bm{x}_{-i}$, all nodes except $i$. In an MRF, $\bm{x}_{-i}$ can  be simplified to the Markov blanket of $i$. Since $\mathbb{P}_{\Phi}(x_i|\bm{x}_{-i})$ is just the product of several factors, if we apply the Gumbel-Softmax trick, the sampling process is differentiable with respect to these factors. A graphical illustration of the overall procedure is shown in Fig. \ref{fig:learning}.

After a burn-in period with $n_{\text{burnin}}$ samples, Gibbs sampling will converge to the stationary distribution. The length of burn-in period is theoretically unknown and is often decided empirically. Gibbs sampling can be expensive because every time we update $\Phi$, we will have a new $\mathbb{P}_{\Phi}$. Therefore, we need to re-enter the burn-in period just to draw $n_{\text{mc}}$ samples where $n_{\text{mc}} \ll n_{\text{burnin}}$, and then update $\Phi$ again. In order to mitigate this problem, we propose a Long-Short-Burnin-Scheme (LSBS). Specifically, at the beginning of each epoch, we run a long burn-in period $n_{\text{long}}$, but at each iteration within that epoch, we only run a short burn-in period $n_{\text{short}}$. We can assume that $\mathbb{P}_{\Phi^t} \approx \mathbb{P}_{\Phi^{t+1}}$ since $\Phi$ will only change by a small amount. Starting from a sample $\bm{x}^t \sim \mathbb{P}_{\Phi^t}$, we can quickly transit to a sample $\bm{x}^{t+1} \sim \mathbb{P}_{\Phi^{t+1}}$ without running a long burnin period. As a result, as opposed to $n_{\text{long}} + n_{\text{mc}}$, we only need to draw $n_{\text{short}} + n_{\text{mc}}$ samples at each iteration, where $n_{\text{mc}} \approx n_{\text{short}} \ll n_{\text{long}}$.

\section{Experiments}
\subsection{Datasets}
We choose five semantic segmentation datasets with diverse contents. Specifically, DeepGlobe challenge \cite{DeepGlobeDemirCVPRWorkshop2018} provides three aerial image datasets: Land, Road and Building. Land is a multi-class (urban, agriculture, rangeland, forest, water and barren) segmentation dataset and it contains 803 satellite images focusing on rural areas. Road is a binary segmentation dataset and it consists of 6226 images captured over Thailand, Indonesia, and India. Building is also a binary segmentation task with 10593 images taken from Las Vegas, Paris, Shanghai, and Khartoum. 

CHAOS \cite{CHAOSKavurMIA2021} is medical image dataset including both computed tomography (CT) and magnetic resonance imaging (MRI) scans of abdomen organs (liver, right kidney, left kidney and spleen). We only use MRI scans, and it has 20 cases and 1270 2D-slices. PROMISE \cite{PROMISE12LitjensMIA2014} contains prostate MRI images, and it has 50 cases and 1377 2D-slices. For all datasets, only training sets are available. We split 60\%/20\%/20\% of the training set as train/val/test set. For simplicity, we resize all images to $256\times 256$.

\subsection{Search Space}
We use UNet \cite{U-NetRonnebergerMICCAI2015} as the backbone. Generally speaking, UNet and other encoder-decoder networks usually contain three types of operations: Normal, Down and Up. We search for both size and width of the convolution kernel and summarize the search space in Table~\ref{tb:search-space}. For the original UNet that has 26 layers, our search space has about $4 \times 10^{23}$ configurations in total.

For a fair comparison, for manually designed architectures, e.g.\ UNet \cite{U-NetRonnebergerMICCAI2015}, UNet++ \cite{UNet++ZhouTMI2020} and BiO-Net \cite{BiO-NetXiangMICCAI2020}, we use their templates and implement the Normal/Up/Down operations in the same way as our search space, but fix the kernel size to be 3. For automatically found architectures, e.g.\ NAS-UNet \cite{NAS-UnetWengAccess2019}, MS-NAS \cite{MS-NASYanMICCAI2020}, and BiX-Net \cite{BiX-NASWangMICCAI2021}, we do not make any modification and use their implementations directly. However, there exist discrepancies. For instance, we use transposed convolution for up-sampling while NAS-UNet \cite{NAS-UnetWengAccess2019} uses dilated transposed convolution and BiX-Net \cite{BiX-NASWangMICCAI2021} uses bilinear interpolation.

\begin{table}[h]
\centering
\caption{\label{tb:search-space}Search space with UNet as the backbone.}
\begin{tabular}{c|c|c}
\hlineB{2}
      Type & Size                           & Width \\ \hline
Normal & 3, 5  & 0.5, 0.75, 1.0, 1.25, 1.5          \\
Down   & 3          & 0.5, 0.75, 1.0, 1.25, 1.5          \\
Up     & 2 & 0.5, 0.75, 1.0, 1.25, 1.5           \\ \hlineB{2}
\end{tabular}
\end{table}

\subsection{Implementation Details} 
In the search phase, we train a super-network using the sandwich rule \cite{UniversallyYuICCV2019} for $T=50$ epochs. Initially, factors are not updated until the super-network is trained for a warmup period of $10$ epochs. The learning rate of network weights starts from 0.0005 and is then decreased by a factor of $(1-\frac{t}{T})^{0.9}$ at each epoch $t$. We use the Adam optimizer with weight decay of 0.0001. The learning rate of MRF factors is fixed at 0.0003 and we also use the Adam optimizer with the same weight decay. For the sampling, we use $n_{\text{long}}=10000$, $n_\text{{short}}=10$ and $n_{\text{mc}}=1$. The temperature parameter $\tau$ in Gumbel-Softmax is fixed at $1$. For inference, we choose $m=5$ and $L=10$ for diverse M-best inference, and the number of binary search iterations is $n_\text{{iter}}=20$. For simplicity, we search on Deepglobe Land \cite{DeepGlobeDemirCVPRWorkshop2018} and the found architectures are evaluated on other datasets. Our results can be improved by searching on each dataset individually. In the re-train phase, we use the same hyper-parameters as in the search phase, except that we train for $T=100$ epochs. We use the same hyper-parameters for both architectures found by our methods as well as the baselines. 

\subsection{Computational Cost}
The overhead of our method comes from Gibbs sampling and inference over a complex loopy graph. Since we run a long burn-in period $n_\text{{long}}$ only at the start of each epoch, and a short burn-in period $n_\text{{short}}+n_\text{mc}$ at each training iteration, the cost of sampling is negligible compared with a forward-backward pass of the neural network. We will discuss the cost of loopy inference algorithms in section \ref{sec:inference}, and since we use approximate inference algorithm as a default, the cost of inference is also minimal. In our experiments, the overhead takes up less than 2\% of the total search time.

\subsection{MRF-UNets Architecture}
MRF-UNet shows several interesting characteristics that differ from the original UNet: (i) it has a larger encoder but a smaller decoder (ii) layers that are connected by the long skip connections are shallower (iii) layers that need to process these concatenated feature maps are wider and also have larger kernel size. As a result, there exists a bottleneck pattern in the encoder and an inverted bottleneck pattern in the decoder. Our observations show that the encoder and decoder do not need to be balanced as in the original UNet and many other encoder-decoder architectures \cite{3DU-NetCicekMICCAI2016,V-NetMilletari3DV2016,LinkNetChaurasiaVCIP2017}. They also demonstrate that the widely adopted ``half resolution, double width'' principle might be sub-optimal in an encoder-decoder network. Indeed, feature maps are concatenated by the long skip connections and are processed by the following layer, which form the most computationally extensive part in the whole network. Therefore, their widths should be smaller to reduce complexity. However, layers that need to process this rich information should be wider and have larger receptive fields.

\begin{figure}[h]
\centering
\includegraphics[width=1\linewidth]{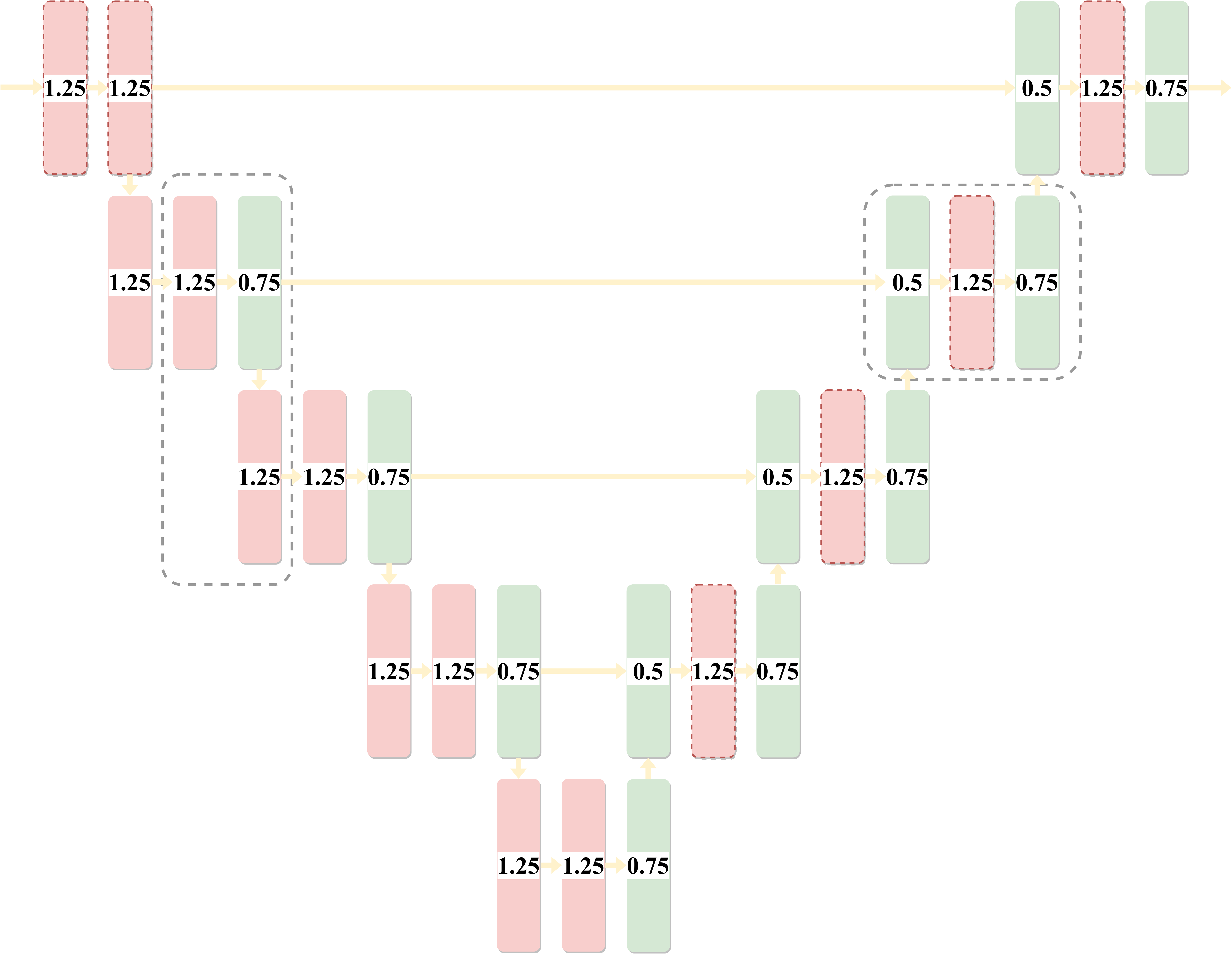}
\caption{\label{fig:mrfunetv2}Architecture of MRF-UNetV2. Numbers inside rectangles are width ratios to the original UNet. Rectangles surrounded by colored dashed lines use $5\times5$ kernels while others use $3\times3$ kernels. The gray dashed line on the left highlights an example of a bottleneck block in the encoder, and the gray dashed line on the right shows an example of an inverted bottleneck block in the decoder.}
\end{figure}

Inspired by these observations, we propose MRF-UNetV2 to emphasize these characteristics. As shown in Fig. \ref{fig:mrfunetv2}, MRF-UNetV2 has a simpler architecture that is easier for implementation, and we show that it can sometimes outperform MRF-UNet in Table \ref{tb:aerialresult} and Table \ref{tb:medicalresult}. We note that a recent trend in NAS is to design a more and more complex search space to include as many candidates as possible \cite{OnWanICLR2022}, but it becomes difficult to interpret the search results. We hope that our observations can inspire practitioners when designing other encoder-decoder architectures.

\subsection{Main Results}
Our benchmarks include manually designed architectures: UNet \cite{U-NetRonnebergerMICCAI2015}, UNet++ \cite{UNet++ZhouTMI2020}, BiO-Net \cite{BiO-NetXiangMICCAI2020} and architectures found with NAS: NAS-UNet \cite{NAS-UnetWengAccess2019}, MS-NAS \cite{MS-NASYanMICCAI2020}, BiX-Net \cite{BiX-NASWangMICCAI2021}. The main results are in Table~\ref{tb:aerialresult} and Table~\ref{tb:medicalresult}. We set the latency target to 1.70ms which is the same as the original UNet, and we also include FLOPs for comparison. Our MRF-UNets outperform benchmarks over five datasets with diverse semantics, while require less computational resources.

\begin{table}[h]
\centering
\caption{\label{tb:aerialresult}mIoU (\%) on DeepGlobe challenge. mean$\pm$std are computed over 5 runs.}
\begin{tabular}{cccccc}
\hlineB{2}
Model      & Land (\%) & Road (\%) & Building (\%) & FLOPs (G)  & Latency (ms) \\ \hline
UNet      & $58.41 \pm 0.52$ & $57.05 \pm 0.13$  & $75.12 \pm 0.09$ & 4.84  & 1.70    \\
UNet++    & $59.53 \pm 0.21$ & $56.96 \pm 0.08$  & $74.27 \pm 0.21$ & 11.76  & 5.11    \\
BiO-Net  & $58.10 \pm 0.37$ & $57.06 \pm 0.22$   &$74.29 \pm 0.15$  & 37.22 & 3.28   \\ \hline
NAS-UNet   & $58.11 \pm 0.91$ & $57.73 \pm 0.56$   & $74.46 \pm 0.19$ & 30.44 & 4.25    \\
MS-NAS   & $58.75 \pm 0.68$ & $57.34 \pm 0.35$   & $74.61 \pm 0.17$ & 24.28 & 3.96    \\
BiX-Net   & $57.96 \pm 0.94$ & $57.74 \pm 0.08$   & $74.63 \pm 0.12$ & 13.28  & 1.87    \\ \hline
MRF-UNet   & \textbf{59.64}$\pm$0.44 & $57.81 \pm 0.17$   & $75.50 \pm 0.09$ & 4.74  & 1.68    \\
MRF-UNetV2 & $58.56 \pm 0.25$ & \textbf{57.90}$\pm$0.23 & \textbf{75.84}$\pm$0.13 & 4.66  & 1.70  \\ 
\hlineB{2}
\end{tabular}
\end{table}

\begin{table}[h]
\centering
\caption{\label{tb:medicalresult}Dice scores (\%) on CHAOS and PROMISE. mean$\pm$std are computed over 5 runs.}
\begin{tabular}{ccccc}
\hlineB{2}
Model      & CHAOS (\%) & PROMISE (\%) & FLOPs (G)  & Latency (ms) \\ \hline
UNet       & $91.16 \pm 0.23$ & $84.60\pm 0.68$   & 4.84  & 1.70    \\
UNet++     & $91.46 \pm 0.15$ & $86.29 \pm 0.35$   & 11.76  & 5.11    \\
BiO-Net  & $91.80 \pm 0.42$ & $86.04 \pm 0.77$   & 37.22 & 3.28    \\ \hline
NAS-UNet & $91.30 \pm 0.65$ & $85.04 \pm 0.90$   & 30.44 & 4.25    \\
MS-NAS    & $91.47 \pm 0.35$ & $85.42 \pm 0.72$   & 24.28 & 3.96    \\
BiX-Net   & $91.22 \pm 0.39$ & $84.35 \pm 0.91$   & 13.28  & 1.87    \\ \hline
MRF-UNet   & $92.03 \pm 0.31$ & \textbf{86.76}$\pm$ 0.32 & 4.74  & 1.68    \\
MRF-UNetV2 & \textbf{92.14}$\pm$0.24 & $86.61 \pm 0.36$   & 4.66  & 1.70  \\ 
\hlineB{2}
\end{tabular}
\end{table}

\section{Ablation Study}
\subsection{Exact vs. Approximate Loopy Inference} \label{sec:inference}
\begin{figure}[h]
\begin{center}
\includegraphics[width=\linewidth]{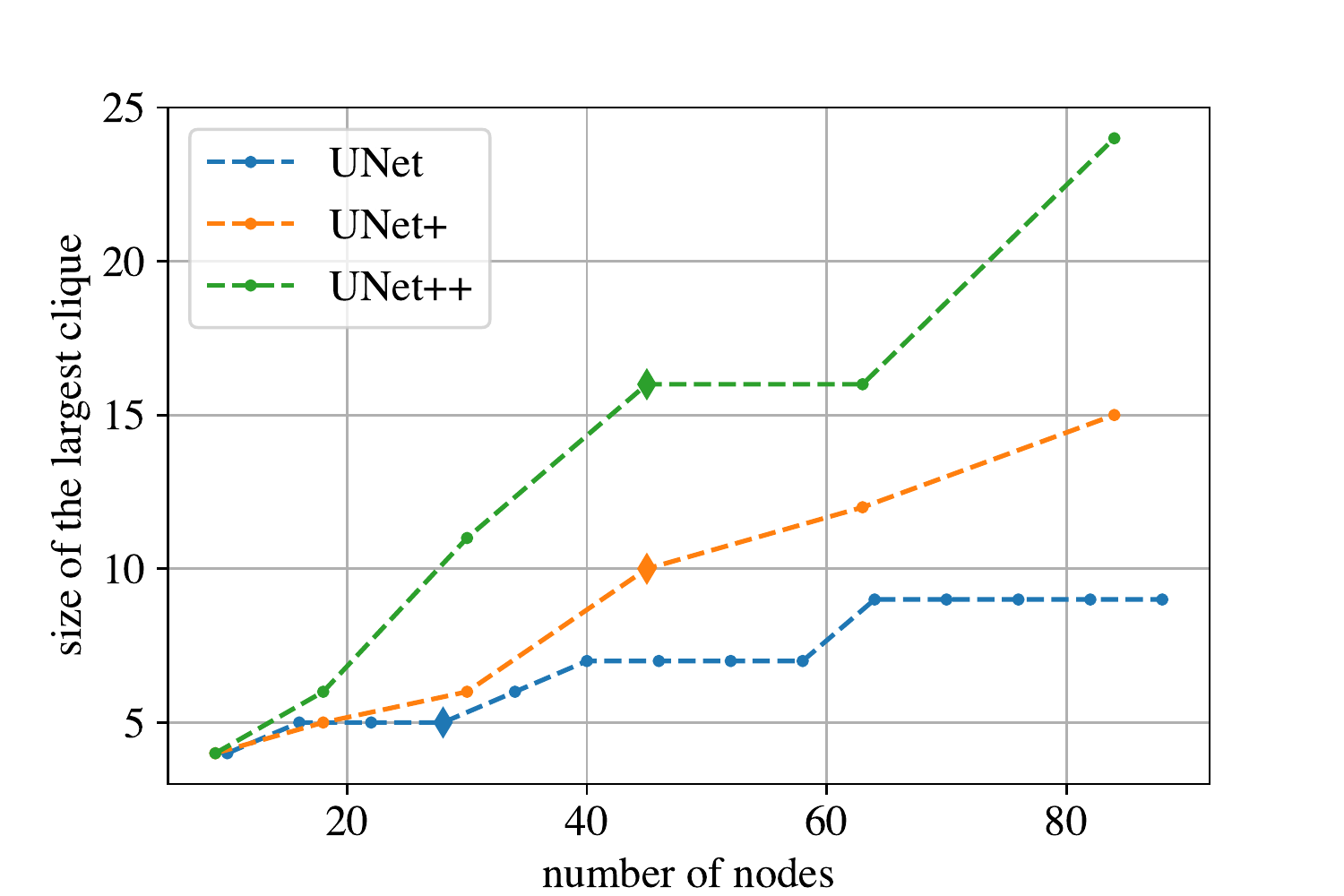}
\caption{\label{fig:clique}Size of the largest clique vs.\ the number of nodes for UNet \cite{U-NetRonnebergerMICCAI2015}, UNet+ \cite{UNet++ZhouTMI2020} and UNet++ \cite{UNet++ZhouTMI2020}. Diamonds indicate the original architectures whose depth is 5.}
\end{center}
\end{figure}

Without our loopy inference extension, AOWS fails on more complex loopy graphs. However, inference on a loopy MRF is a NP-hard problem \cite{PGMKoller2009}, so we cannot always hope for exact solutions. Here we use Max-Product Clique Tree algorithm (MPCT) for exact solutions, and Max-Product Linear Programming (MPLP) for approximate inference \cite{PGMKoller2009}. We compare architectures found by MPCT and MPLP in Table \ref{tb:inf}. They usually obtain very similar solutions and their results are almost identical. 
The complexity of MPCT and in general of exact inference is $O(|X|^{|C|})$ where $|X|$ is the cardinality of factors and in our experiments it is 10, and $|C|$ is the size of the largest clique. Generally, $|C|$ increases when the graph is more densely connected, and in the worst case $|C| = n$ the number of nodes when the graph is fully connected, e.g.\ DenseNet \cite{DenseNetHuangCVPR2017}. In Fig. \ref{fig:clique}, we show the size of the largest clique vs.\ the number nodes for UNet \cite{U-NetRonnebergerMICCAI2015}, UNet+ \cite{UNet++ZhouTMI2020} and UNet++ \cite{UNet++ZhouTMI2020}. UNet++, being more densely connected, has a much larger clique size than UNet. The clique size of the original UNet is 5, and MPCT already takes several minutes on our MacBook Pro (14-inch, 2021). Note that we need to repeat the inference for $m \times n_\text{{iter}}=100$ times, and it will soon become infeasible if we want to apply MPCT on deeper UNet or UNet+/UNet++. Nevertheless, MPLP can converge within a few seconds. Since they usually find similar solutions, we use MPLP as a default.

\begin{table}[h]
\centering
\caption{\label{tb:inf}Evaluating architectures found by MPCT and MPLP on DeepGlobe Land. mean$\pm$std are computed over 5 runs.}
\begin{tabular}{ccc}
\hlineB{2}
Algorithm & MPCT & MPLP \\ \hline
mIoU (\%) & $59.56\pm0.57$ & \textbf{59.64}$\pm$0.44   \\ \hlineB{2}
\end{tabular}
\end{table}

\subsection{Diverse Solutions}
As shown in \cite{NASYangICLR2020,EvaluatingYuICLR2020}, there exists an inconsistency between the true rank of an architecture and its rank on a weight-sharing network, but we can still find architectures that are reasonably good by examining other top solutions. This motivates us to apply diverse M-best inference \cite{DiverseMBestBatraECCV2012}. In Table~\ref{tb:diverse5}, we show the results of diverse 5-best. The MAP solution is not the best quality and diverse M-best inference can greatly improve our results. It also helps reduce the variance in the search phase since we can evaluate $m$ highly probable solutions at the same time, and the total computational cost is thus decreased from $m \times (\text{cost}_{\text{search}} + \text{cost}_{\text{eval}})$ to $\text{cost}_{\text{search}} + m \times \text{cost}_{\text{eval}}$.

Calibration \cite{CalibrationGuoICML2017} is critical for medical diagnosis and deep ensembles \cite{DeepEnsemblesLakshminarayananNeurIPS2017} have been shown to effectively reduce the calibration error. Compared with a deep ensemble that consists of the same architecture from different initialization, we can form an ensemble with a diverse set of solutions. As shown in Table \ref{tb:ensemble}, our diverse ensemble achieves a lower Expected Calibration Error (ECE) on CHAOS.

Should we further increase $m$ if it is so helpful? The answer is no: further increasing $m$ generally does not help us to find a better solution, and the majority of solutions are often of sub-optimal performance so it does not help reduce the variance. In Table \ref{tb:diverse10}, we show the result of diverse 10-best. We choose a higher $L=20$ and expect that the best solution will come later, since it leads to a lower dissimilarity target in Eq. (\ref{eq:L}). Indeed, the best architecture is now the 5th one instead of the 3rd, but it does not show a better performance and many solutions are of similar quality. Therefore, we do not benefit from increasing $m$, while it adds the cost of both inference and evaluation.

\begin{table}
\centering
\caption{\label{tb:diverse5}Evaluating architectures found by diverse 5-best on DeepGlobe Land ($L=10$). mean$\pm$std are computed over 5 runs.}
\begin{tabular}{cccccc}
\hlineB{2}
Solution & 1 & 2 & 3 & 4 & 5 \\ \hline
mIoU (\%) & 57.37$\pm$0.61  & 57.41$\pm$0.32  & \textbf{59.64}$\pm$0.44  & 59.43$\pm$0.46  & 56.59$\pm$0.70  \\ \hlineB{2}
\end{tabular}

\vspace*{0.25cm}

\caption{\label{tb:diverse10}Evaluating architectures found by diverse 10-best on DeepGlobe Land ($L=20$). mean$\pm$std are computed over 5 runs.}
\begin{tabular}{cccccc}
\hlineB{2}
Solution & 1 & 2 & 3 & 4 & 5 \\ \hline
mIoU (\%) & 57.02$\pm$0.53 & 58.37$\pm$0.41  & 56.94$\pm$0.55  & 59.37$\pm$0.41  & \textbf{59.56}$\pm$0.22  \\ \hline
Solution & 6 & 7 & 8 & 9 & 10 \\ \hline
mIoU (\%) & 57.86$\pm$0.41  & 59.45$\pm$0.92& 57.82$\pm$0.51  & 57.79$\pm$0.48  & 57.84$\pm$0.39 \\
\hlineB{2}
\end{tabular}

\vspace*{0.25cm}

\caption{\label{tb:ensemble}Comparing deep ensemble \cite{DeepEnsemblesLakshminarayananNeurIPS2017} with our diverse deep ensemble on CHAOS. Lower is better. mean$\pm$std are computed over 5 runs.}
\begin{tabular}{ccc}
\hlineB{2}
Model & Deep Ensemble & Diverse Deep Ensemble \\ \hline
ECE (\%)     & $0.7091\pm0.0140$  & \textbf{0.6872}$\pm$0.0126  \\ \hlineB{2}
\end{tabular}
\end{table}

\subsection{Pairwise Formulation and Differentiable Parameter Learning}
Except diverse M-best loopy inference, we make other two modifications: pairwise formulation and differentiable parameter learning. In table \ref{tb:mrf}, we compare the architectures found by our our MRF-NAS vs. MRF-NAS without pairwise factors and MRF-NAS without differentiable parameter learning. 

\begin{table}[h]
\centering
\caption{\label{tb:mrf}Evaluating architectures found by A1: MRF-NAS w/o pairwise factors, A2: MRF-NAS w/o differentiable parameter learning and MRF-UNet on DeepGlobe Land. mean$\pm$std are computed over 5 runs.}
\begin{tabular}{cccc}
\hlineB{2}
Architecture & A1 & A2  & MRF-UNet \\ \hline
mIoU (\%)     &  59.17$\pm$0.58 & 59.33$\pm$0.31 & \textbf{59.64}$\pm$0.44  \\ \hlineB{2}
\end{tabular}
\end{table}

\section{Conclusion}
In this paper, we propose MRF-NAS that extends and improves AOWS \cite{AOWSBermanCVPR2020} with a more general framework based on a pairwise Markov Random Field (MRF) formulation, which leads to applying various statistical techniques for MAP optimization. With diverse M-best loopy inference algorithms and differentiable parameter learning, we find an architecture, MRF-UNet, with several interesting characteristics. Through the lens of these characteristics, we identify the sub-optimality of the original UNet and propose MRF-UNetV2 with a simpler architecture that can further improve our results. MRF-UNets, albeit requiring less computational resources, outperform several SOTA benchmarks over three aerial image datasets and two medical image datasets that contain diverse contents. This demonstrates that the found architectures are robust and effective.

\section*{Acknowledgements}
We acknowledge support from the Research Foundation - Flanders (FWO) through project numbers G0A1319N and S001421N, and funding from the Flemish Government under the Onderzoeksprogramma Artifici\"{e}le Intelligentie (AI) Vlaanderen programme.

\section*{Appendix}
\begin{figure}[h]
\centering
\includegraphics[width=1\linewidth]{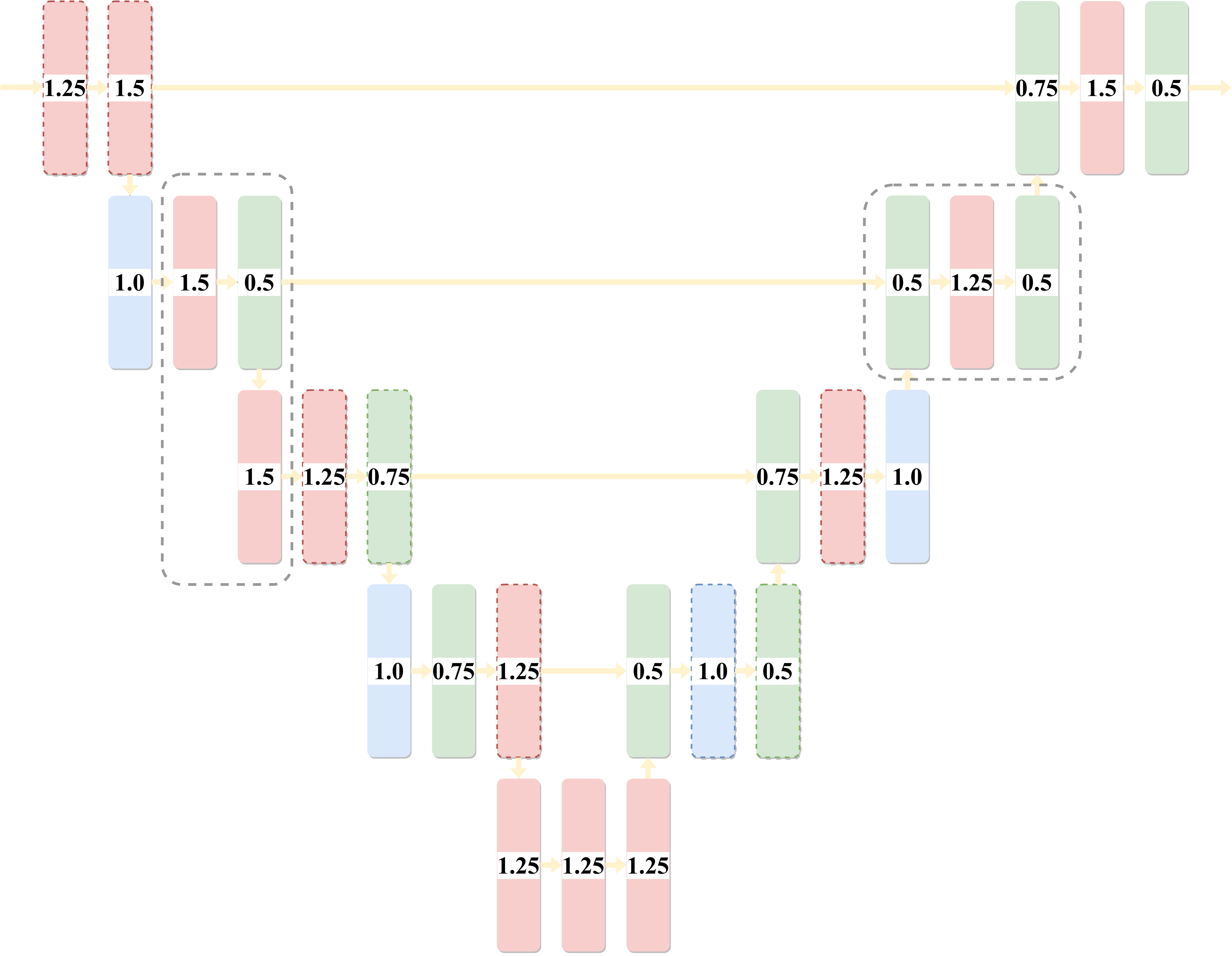}
\caption{\label{fig:mrfunet}Architecture of MRF-UNet. Numbers inside rectangles are width ratios to the original UNet. Rectangles surrounded by colored dashed lines use $5\times5$ kernels while others use $3\times3$ kernels. The gray dashed line on the left highlights an example of a bottleneck block in the encoder, and the gray dashed line on the right shows an example of an inverted bottleneck block in the decoder.}
\end{figure}

The architecture of MRF-UNet is shown in Fig. \ref{fig:mrfunet}.

\end{document}